\documentclass[10pt,journal,compsoc]{IEEEtran}
\usepackage{graphicx}
\usepackage{booktabs}
\usepackage{amsmath,amssymb}
\usepackage{algorithm}
\usepackage[noend]{algpseudocode}
\usepackage{url}

\usepackage{cite}

\usepackage[colorlinks,linkcolor=blue,citecolor=blue]{hyperref}
\usepackage[capitalize]{cleveref}   
\crefname{table}{Table}{Tables}
\crefname{figure}{Fig.}{Figs.}
\crefname{section}{Section}{Sections}
\crefname{equation}{Eq.}{Eqs.}

\newcommand{\valAnBaseExpr}{0.507}
\newcommand{\valAnCircExpr}{0.560}
\newcommand{\valAnExprDelta}{+0.052}
\newcommand{\valRafDelta}{-0.005}
\newcommand{\valAngBase}{0.107}
\newcommand{\valAngCirc}{0.079}
\newcommand{\valAngAU}{0.078}
\newcommand{\valRsaFsfmFrozen}{+0.071}

\newcommand{\valRsaFarlFrozen}{+0.233}
\newcommand{\valIsoAbaw}{+0.548}
\newcommand{\valIsoAffectnet}{+0.667}
\newcommand{\valIsoCombinedP}{0.014}
\newcommand{\valIsoAbawPtwo}{0.171}
\newcommand{\valIsoAnPtwo}{0.083}
\newcommand{\valHolmReject}{0}
\newcommand{\valHolmFamily}{4}
\newcommand{\valCiAbSevMean}{-0.008}
\newcommand{\valCiAbSevCI}{[-0.009, -0.007]}
\newcommand{\valCiAbScoreMean}{+0.000}
\newcommand{\valCiAbScoreCI}{[-0.001, +0.002]}

\newcommand{\valRsaBaseMean}{+0.164}
\newcommand{\valRsaBaseStd}{0.009}
\newcommand{\valRsaCircMean}{+0.177}
\newcommand{\valRsaCircStd}{0.006}
\newcommand{\valRsaShiftP}{0.062}
\newcommand{\valRsaSiglip}{+0.369}
\newcommand{\valDistLinPmtl}{1.228}
\newcommand{\valDistInstPmtl}{1.232}
\newcommand{\valDistInstP}{0.094}
\newcommand{\valDistGwPmtl}{1.216}
\newcommand{\valCtrlUniPmtl}{1.227}
\newcommand{\valCtrlCircUniP}{0.625}
\newcommand{\valCtrlAnUniExpr}{0.565}
\newcommand{\valCtrlAnUniVsBase}{+0.057}
\newcommand{\valCtrlAnLinVsUniP}{0.031}
\newcommand{\valEsAbawSevP}{0.031}
\newcommand{\valEsAbawQuadP}{0.031}
\newcommand{\valEsAnSevP}{0.094}
\newcommand{\valStabMaxCentSE}{0.020}
\newcommand{\valStabMinDist}{0.054}
\newcommand{\valStabMinDistSE}{0.007}
\newcommand{\valStabBoot}{1000}

\begin{document}
\title{The Circumplex Degeneracy Behind the Rare-Class Limit in Affect Recognition}

\author{Van Thong~Huynh, Hong Hai~Nguyen,~
        and~Soo-Hyung~Kim
\IEEEcompsocitemizethanks{\IEEEcompsocthanksitem Van Thong Huynh is with Faculty of CSE, Ho Chi Minh City University of Technology (HCMUT), VNUHCM, Vietnam. Hong Hai Nguyen is with Dept. of AI, FPT University, Vietnam. Soo-Hyung Kim is with Dept. of AI Convergence, Chonnam National University,
South Korea.\protect\\
\IEEEcompsocthanksitem Corresponding author: Van Thong~Huynh.}

}

\markboth{}
{Huynh \MakeLowercase{\textit{et al.}}: The Circumplex Degeneracy Behind the Rare-Class Limit}

\IEEEtitleabstractindextext{%
\begin{abstract}
In-the-wild expression recognition persistently fails on a few rare emotions, and the
standard explanation is class imbalance. Through a controlled multi-task study on two
benchmarks, we show the failure is instead a property of affect geometry: the rare classes
are degenerate on Russell's circumplex, and that degeneracy bounds what any loss or cost
can achieve. Our instrument is a circumplex-cost optimal-transport term that prices
expression confusions by their valence--arousal distance. The term improves the official
score and expression macro-F1, but a control most studies omit shows the gain is not
geometric: a uniform cost, equivalent to a generic confidence penalty, matches it on
Aff-Wild2 ($p=\valCtrlCircUniP{}$) and significantly exceeds it on AffectNet
($\valCtrlAnUniVsBase{}$ over base, larger than the circumplex). What the geometry reshapes
is the structure of the errors, making them affectively nearer the truth on Aff-Wild2
($p=\valEsAbawSevP{}$ against the uniform control), an effect that does not survive on
AffectNet, where a visual confound at the far corner of the circumplex overwhelms it. The
rare-class failure, by contrast, is stable across both datasets we examine: the degenerate
pairs (anger--fear on Aff-Wild2, anger--contempt on AffectNet) resist frequency-based
interventions, the transport term, and an action-unit-augmented cost built specifically to
separate them. We conclude that progress on rare expressions requires representations that
distinguish the classes, not supervision that reprices their confusions, and we provide the
controls and metrics needed to tell the two apart.
\end{abstract}

}

\maketitle

\IEEEdisplaynontitleabstractindextext
\IEEEpeerreviewmaketitle

\section{Introduction}
\label{sec:intro}
In-the-wild expression recognition reliably reaches usable accuracy on
common emotions and reliably fails on a few rare ones: anger and fear sit near zero F1 in
multi-task affect systems, year after year. Since the categorical task is long-tailed and
its metric, macro-averaged F1, gives the rarest classes as much weight as the most common
ones, the standard interpretation is that the failure is a problem of class frequency, to be fixed
by re-weighting, balanced sampling, focal losses, margin losses~\cite{cao2019ldam}, or logit
adjustment~\cite{menon2021logit}. These methods reshape the loss by label counts, and on the
hardest classes they do not work. Our study examines why, and answers with a different
diagnosis: the failure is not about how often these classes occur but about where they sit in
affect space.

Russell's circumplex model~\cite{russell1980circumplex} places emotions on a valence--arousal
plane, and multi-task affect datasets annotate that plane directly, since expression labels
co-occur with valence--arousal labels on overlapping frames. Estimated from the data, the
circumplex reveals that the classes that fail are not merely rare but geometrically
degenerate: anger and fear nearly coincide on Aff-Wild2, and anger and contempt nearly
coincide on AffectNet. We argue that this degeneracy, rather than frequency, bounds what any
loss or cost can achieve on these classes, and we establish the claim with controls rather
than assertion.

Our instrument is a circumplex-cost optimal-transport (OT) term that prices expression
confusions by their valence--arousal distance, so that confusing affectively distant emotions
costs more than confusing affectively near ones. The term is a natural way to put the
geometry into training, and it does improve the official multi-task score and expression
macro-F1. The paper's main methodological finding is that this improvement is not
what it appears to be. A control that this literature almost never runs, replacing the
circumplex cost with a uniform cost, reproduces the entire gain: the uniform cost is
statistically indistinguishable from the true circumplex on Aff-Wild2
($p=\valCtrlCircUniP{}$) and significantly better on AffectNet
($\valCtrlAnUniVsBase{}$ over the base recipe, exceeding the circumplex). Given that a uniform
cost reduces the OT term to a generic true-class confidence penalty, the score gain is an
effect of adding any non-trivial penalty, not of affect geometry. The geometry does
change the structure of the model's errors, an effect macro-F1 cannot see: on Aff-Wild2 it
makes the surviving mistakes affectively nearer the truth ($p=\valEsAbawSevP{}$ against the
uniform control), but this does not survive on AffectNet, where a visually driven
contempt--happiness confusion at the far corner of the circumplex overwhelms it.

Against these qualified and dataset-specific effects, the rare-class failure itself is
stable. The degenerate pairs resist every intervention we try: frequency-based
losses, the transport term, and an action-unit-augmented cost we build
specifically to separate anger from fear, which raises their cost from a near-degenerate
value to a well-resolved one and still does not recover anger. The diagnosis that survives
all of these controls is representational: a cost can reweight confusion among classes the
features already distinguish, but it cannot manufacture separability that the representation
lacks. Progress on rare expressions must therefore come from features that tell the classes
apart, not from supervision that reprices their confusions. Our contributions are as follows.
\begin{itemize}
\item A controlled isolation of what affect-geometry supervision actually does
(\cref{sec:results}). A circumplex-cost OT term raises the multi-task score, but shuffled and
uniform-cost controls, rarely run in this literature, show the improvement is a generic
confidence-penalty effect rather than a geometric one, on two benchmarks.
\item A diagnosis of the rare-class limit as a circumplex degeneracy that is representational,
not frequency- or cost-driven (\cref{sec:degeneracy}). It holds across two datasets and is
sharpened by an action-unit-routing experiment in which a cost geometry built to separate the
degenerate pair still fails to recover it.
\item Metrics and controls for telling geometric error-shaping apart from generic
regularization, namely error severity, quadrant preservation, and the shuffled-cost control,
with an honest account of where a geometric error effect appears and where it does not
(\cref{sec:results:errstruct}).
\item A controlled study of the representation, comparing pretraining domains and probing,
with representational-similarity analysis, when affect geometry is already encoded by the
features (\cref{sec:results:backbone,sec:degeneracy:rsa}).
\end{itemize}

\section{Related Work}
\label{sec:related}

\subsection{Structured Label Costs and Wasserstein Training}
Frogner et al.~\cite{frogner2015wasserstein} introduced learning with a Wasserstein loss
under a prior ground metric between classes, showing that exploiting semantic structure in
the label space improves learning when classes are related. Our term is the linear special
case of this family, with the ground metric instantiated by affective rather than semantic
structure and estimated from the task's own auxiliary annotations rather than from an
external ontology. Optimal transport has also been used to refine labels under partial
supervision and imbalance: SoLar~\cite{wang2022solar} shapes pseudo-labels with a Sinkhorn
objective in partial-label learning. These works use OT machinery for assignment; we use it
as a geometric penalty. Closest in form are EMD-style classification losses with structured
ground metrics: the squared-EMD loss for ordinal and related classes~\cite{hou2016emd}, and,
for emotion specifically, a hierarchical EMD over a taxonomy of text emotion
labels~\cite{eemd2025}. The ground metrics in this line are taxonomic or ordinal, and the
dimensional geometry of affect estimated from the task's own valence--arousal annotations
offers a natural instance for facial affect. We use that instance not as a proposed method
but as an instrument: a structured cost whose effect we can isolate against geometry-free
controls. That control is what most separates our study from this line, where structured-cost
losses are reported to help without a baseline that strips the geometry out, leaving open
whether the gain is geometric or generic, the question this paper answers.

\subsection{Long-Tailed Recognition}
The standard treatments of class imbalance modify the loss by label frequency: re-weighting,
class-balanced sampling, margin enlargement for tail classes (LDAM~\cite{cao2019ldam}),
asymmetric losses for multi-label tails, and logit adjustment~\cite{menon2021logit}. All are
geometry-blind: they encode how often a class occurs, not where it lies relative to others.
Optimal transport has also been brought to bear on imbalance: an OT view of class-imbalanced
recognition transports the predicted label distribution toward a balanced reference to
counter the skew~\cite{jin2023otimbalance}. That line still operates on frequency, moving
mass to equalize counts, whereas our cost moves mass by inter-class affect distance and is
indifferent to how often a class occurs. Our experiments include the frequency-based methods
on the identical pipeline; none improves the overall score, and, like every cost we try,
none resolves the degenerate anger--fear pair. We do not position the circumplex term as a
better long-tailed loss: our controls show its score advantage over these methods is the
effect of a generic confidence penalty rather than of affect geometry, and what it shares
with them is the inability to separate the classes the representation collapses.

\subsection{Circumplex- and Language-Guided Affect Recognition}
The circumplex model~\cite{russell1980circumplex} underlies the valence--arousal annotation
of modern affect datasets, and prior work has exploited the EXPR--VA relationship for
recognition. CAGE~\cite{wagner2024cage} trains expression inference jointly with
valence--arousal targets and shows consistent gains, treating VA as an auxiliary regression
whose information reaches the classifier through shared features. We differ in mechanism: the
circumplex enters through the loss geometry of the categorical task itself, as a cost on
confusions. Label-distribution learning makes a related use of the geometry on the target
side: Le et al.~\cite{le2023ldl} construct soft expression targets from neighborhood
structure in the valence--arousal space. Our term instead leaves the targets intact and
re-prices the posterior's errors, a cost-side use of the same geometry. The distinction
matters empirically: our ablation implements the natural prediction-level coupling and finds
it harmful, whereas the cost-side term improves the score, though our controls later show
that improvement is not attributable to the geometry.

A parallel and increasingly prominent way to inject affect structure is through natural
language. Vision--language models supply emotion knowledge that purely visual training lacks:
EmoCLIP aligns video with affective text descriptions for zero-shot expression
recognition~\cite{foteinopoulou2024emoclip}, and Exp-CLIP transfers semantic knowledge from
large language models into the expression classifier~\cite{zhao2025expclip}. These methods
import similarity structure from a language model trained on emotion-laden text; our term
instead estimates that structure directly from the valence--arousal annotations the dataset
already carries, with no external model. The two views meet in our representational analysis
(\cref{sec:degeneracy:rsa}): the geometry a vision--language encoder acquires from language is,
empirically, the same circumplex structure our ground cost supplies explicitly, which is why
the term's error-shaping appears on a face self-supervised encoder and is absent on a
vision--language one whose features already encode it.

\subsection{Optimal Transport and Representation Alignment in Face Analysis}
OT machinery has entered facial affect through other doors: matching identities to factor
out subject variation in expression recognition~\cite{jeong2022otfer}, and aligning student
and teacher representations when distilling privileged multimodal
information~\cite{aslam2024distill}. In both, transport couples samples or features; in our
use it prices label-space errors. Separately, our representational-similarity analysis
(\cref{sec:degeneracy:rsa}) follows the tradition of comparing learned feature geometry to a
reference structure~\cite{kriegeskorte2008rsa}; we use it to explain why a vision--language
encoder already encodes the circumplex while a face-SSL encoder does not.

\subsection{Cross-Corpus Generalization in Expression Recognition}
\label{sec:related:crosscorpus}
Expression recognizers transfer poorly across datasets: a model trained on one corpus drops
sharply on another because of differing label conventions, demographics, and capture
conditions, and a body of work addresses this with domain adaptation and generalization,
including a unified cross-domain benchmark with adversarial graph
learning~\cite{chen2022crossdomain}. That line adapts a model's features to a target corpus.
Our use of a second corpus is different and narrower: we test whether the findings about the
circumplex cost, its generic score gain and its error-shaping, hold on a dataset whose
circumplex and whose degenerate pair are estimated independently. They do for the score gain
and the degeneracy, and do not for the error-shaping, which is the cross-dataset evidence on
which the diagnosis rests rather than a claim about domain transfer.

\subsection{Multi-Task Affect, Action Units, and the ABAW Challenge}
Multi-task VA/EXPR/AU systems span shared trunks with task-specific heads, distribution
matching and co-annotation across tasks~\cite{kollias2024distribution}, and the challenge
series' own baselines~\cite{kollias2024abaw7,kollias2026affect}. A recurring theme is that
action units carry information complementary to the categorical label, exploited through
AU relation graphs and cross-task attention~\cite{abaw2022augraph,abaw2024auassisted} and,
more recently, through vision--language joint modeling of AUs and their
descriptions~\cite{ge2024vlfau}. This motivates our action-unit-routing experiment
(\cref{sec:degeneracy:au}), which tests whether a cost built from AU signatures, which
distinguish the degenerate pair in principle, can rescue the classes the circumplex cannot
separate; we route the AU signal into the ground cost rather than into the features
precisely to test whether the limit is one of cost geometry or of representation, and report
that it does not lift the degenerate pair. Our baseline follows the shared-trunk recipe with
a face self-supervised encoder~\cite{wang2025fsfm}, masked per-task losses, and homoscedastic
uncertainty weighting~\cite{kendall2018multi}; we change exactly one thing at a time on top
of it, so that every reported difference is attributable to the change under test.

\section{Method}
\label{sec:method}

\subsection{Problem Setting and Evaluation}
\label{sec:protocol}
The primary benchmark is the multi-task track on the Aff-Wild2 corpus and challenge series~\cite{kollias2026affect,
kollias2025emotions,kollias2025advancements,kollias2024behaviour4all,kollias2024abaw7,
kollias20246th,kollias2024distribution,kollias2023abaw2,kollias2023abaw,kollias2022abaw,
kollias2021analysing,kollias2020analysing,kollias2021affect,
kollias2019expression,kollias2019face,kollias2019deep,zafeiriou2017aff}: video-disjoint
train and validation splits, each
frame carrying up to three annotations (valence and arousal in $[-1,1]$; one of eight
expressions; twelve binary AUs). Invalid entries are marked by sentinels and excluded per
task from both losses and metrics. Only about $37\%$ of training frames carry all three
labels, so partial-label masking is the default condition. The circumplex centroids of
\cref{sec:method:geometry} are computed on the $52{,}229$ frames carrying both an expression
and VA, the largest jointly annotated subset. The official score is
\begin{equation}
P_{\mathrm{MTL}} \;=\; \tfrac{1}{2}\,(\rho_v + \rho_a) \;+\; \tfrac{1}{8}\textstyle\sum_{c}
\mathrm{F1}_c^{\mathrm{EXPR}} \;+\; \tfrac{1}{12}\textstyle\sum_{k} \mathrm{F1}_k^{\mathrm{AU}},
\label{eq:pmtl}
\end{equation}
where $\rho$ is the concordance correlation coefficient. The two additional benchmarks,
\cref{sec:exp:datasets}, are used to test generalization and the dependence on locally
estimable geometry.


\subsection{Base Multi-Task Learning Framework}
A shared encoder $f$ feeds three linear heads: VA (two outputs, tanh-bounded), EXPR
(eight logits), and AU (twelve logits). Let $m^{t}_i \in \{0,1\}$ indicate whether frame $i$
carries a valid label for task $t$; each task's loss is averaged only over its valid frames.
For VA we use a concordance loss per dimension,
\begin{align}
\mathcal{L}_{\mathrm{VA}} &= 1 - \tfrac{1}{2}\big(\mathrm{CCC}(\hat v, v) +
\mathrm{CCC}(\hat a, a)\big), \\
\mathrm{CCC}(\hat y, y) &= \frac{2\,\mathrm{cov}(\hat y, y)}
{\sigma_{\hat y}^2 + \sigma_{y}^2 + (\mu_{\hat y} - \mu_{y})^2},
\end{align}
computed over VA-valid frames in the batch; for EXPR, class-weighted focal cross-entropy
$\mathcal{L}_{\mathrm{focal}} = -w_y (1-p_y)^{\gamma} \log p_y$ with $\gamma{=}2$; and for
AU, binary cross-entropy with per-unit positive re-weighting. The three masked losses are
combined by homoscedastic uncertainty weighting~\cite{kendall2018multi},
$\mathcal{L} = \sum_t \tfrac{1}{2} e^{-s_t} \mathcal{L}_t + \tfrac{1}{2} s_t$ with learned
per-task log-variances $s_t$. The encoder is a face self-supervised
ViT-B/16~\cite{wang2025fsfm}; training uses a one-epoch head warmup with the encoder frozen,
then unfreezes its upper $40\%$ at a reduced learning rate; an exponential moving average of
the weights is used for evaluation. Rare-expression frames are oversampled by inverse class
frequency. The recipe is identical in all conditions; only the EXPR loss changes.

\subsection{Data-Driven Estimation of the Circumplex Geometry}
\label{sec:method:geometry}
For each expression class $c$ we compute a centroid $\mu_c \in \mathbb{R}^2$, the mean
(valence, arousal) over all training frames annotated with both class $c$ and VA. The ground
cost between classes is the Euclidean distance $D_{cc'} = \lVert \mu_c - \mu_{c'} \rVert_2$.
No hand-tuned geometry is involved; the structure (\cref{fig:circumplex}) emerges from the
annotations and reproduces Russell's circumplex~\cite{russell1980circumplex}. On Aff-Wild2 the
maximum pairwise distance is $1.142$ (anger--happiness) and the minimum is $0.054$
(anger--fear); anger has three neighbors within about a sixth of the maximum distance. The
same procedure applied to AffectNet recovers a different but equally interpretable geometry
whose own degenerate pair is anger--contempt ($d=0.064$); this dataset-specific structure is
what the cross-dataset experiments of \cref{sec:results:cross} exploit. The geometry is a
robust property of the data rather than a small-sample artifact: a nonparametric bootstrap
over the jointly-labeled frames (\valStabBoot{} resamples) pins every class centroid to a
radial standard error of at most \valStabMaxCentSE{} against an off-diagonal scale near
$0.60$, including the rarest class, and the degenerate anger--fear distance is stable at
$\valStabMinDist{}\pm\valStabMinDistSE{}$.

\subsection{Circumplex-Cost Optimal-Transport Loss}
Let $p(x) = \mathrm{softmax}(z(x))$ be the predicted expression distribution for a frame with
true class $y$. The term is the expected transport cost of the predicted mass to the
true class:
\begin{equation}
\mathcal{L}_{\mathrm{OT}}(x, y) \;=\; \textstyle\sum_{c} p_c(x)\, D_{c\,y}.
\label{eq:ot}
\end{equation}
This is the linear (assignment-to-a-point) special case of the Wasserstein
loss~\cite{frogner2015wasserstein}: given that the target is a point mass at $y$, the transport
plan is trivial and no Sinkhorn iteration is needed. Mass placed on affect-distant classes is
penalized in proportion to its circumplex distance from the truth, while mass on affect-near
classes is nearly free. The gradient makes the mechanism explicit. With
$p = \mathrm{softmax}(z)$ and $\partial p_c / \partial z_k = p_c(\delta_{ck} - p_k)$,
\begin{equation}
\frac{\partial \mathcal{L}_{\mathrm{OT}}}{\partial z_k}
\;=\; p_k\,\big(D_{k\,y} - \bar{D}_y\big), \qquad
\bar{D}_y = \textstyle\sum_{c} p_c\, D_{c\,y},
\label{eq:otgrad}
\end{equation}
where $\bar{D}_y$ is the posterior-mean cost. Cross-entropy's gradient,
$\partial \mathcal{L}_{\mathrm{CE}} / \partial z_k = p_k - \delta_{ky}$, suppresses every
wrong class in proportion to its probability alone, blind to which class it is.
\Cref{eq:otgrad} instead suppresses a class only if its cost exceeds the current
posterior-average ($D_{k\,y} > \bar{D}_y$) and otherwise raises it: probability mass migrates
from affect-distant classes toward the true class and its cheap neighbors. The full
expression loss is $\mathcal{L}_{\mathrm{EXPR}} = \mathcal{L}_{\mathrm{focal}} + \lambda\,
\mathcal{L}_{\mathrm{OT}}$ with $\lambda = 1$; the term adds no parameters, no inference cost,
and one matrix lookup per batch at training time.

\subsection{Alternative Coupling via Prediction-Level Consistency}
A seemingly natural companion, and the direct formalization of treating VA as auxiliary
supervision for EXPR, is a prediction-level consistency term that pulls predicted VA toward
the predicted expression's expected centroid,
$\mathcal{L}_{\mathrm{cons}} = \lVert \hat{v} - \textstyle\sum_c p_c\,\mu_c \rVert_2^2$,
computed on frames carrying both labels. Our ablation (\cref{sec:results:ablation}) shows
this term degrades VA and the total score, both alone and in combination. The final method is
\cref{eq:ot} alone: the geometry helps as a loss metric on the categorical task, not as a
constraint tying the two heads' predictions together.

\subsection{Distributional Generalizations of the Ground Cost}
\label{sec:method:distributional}
The linear form represents each class by a single centroid $\mu_c$, which discards two kinds
of structure the annotations contain. We formulate two refinements and evaluate them in
\cref{sec:results:distributional}; both are stated here so the comparison is between fully
specified alternatives rather than against an unexplored space.

The first is an instance-adaptive cost. Where a frame carries its own valence--arousal
annotation $a_i$, the cost of placing mass on class $c$ can be anchored at the frame's actual
affect coordinate rather than at the true class's centroid, $\tilde D^{i}_{c} = \lVert \mu_c -
a_i \rVert_2$, falling back to the class-level $D_{c\,y}$ on frames without VA. This makes the
supervision reflect where the particular frame sits on the circumplex, so two frames labeled
with the same expression but lying at different points of its VA distribution are priced
differently. The second is a dispersion-aware cost. Each class has not only a mean but an
empirical valence--arousal covariance $\Sigma_c$ (anger's spread is broad along arousal; the
catch-all class is multi-modal), and treating each class as a Gaussian $\mathcal{N}(\mu_c,
\Sigma_c)$ yields a closed-form $2$-Wasserstein ground cost
\begin{equation}
D^{\mathrm{W}_2}_{cc'} = \lVert \mu_c - \mu_{c'} \rVert_2^2 +
\mathrm{Tr}\!\big(\Sigma_c + \Sigma_{c'} - 2(\Sigma_c^{1/2}\Sigma_{c'}\Sigma_c^{1/2})^{1/2}\big),
\label{eq:w2}
\end{equation}
which lets within-class dispersion inform the cost and could in principle separate two classes
whose means nearly coincide but whose distributions do not. Both refinements substitute their
cost matrix into \cref{eq:ot} and leave the rest of the method unchanged. \Cref{sec:results:distributional} reports that neither improves on the linear form, which is
why the linear cost is the cost we analyze.

\subsection{Geometry Controls and Error-Structure Metrics}
\label{sec:method:controls}
To decide whether any effect of the transport term is due to the circumplex geometry or to
the mere presence of an optimal-transport penalty, we run two controls that keep the term and
change only the ground cost. The permuted control deranges the class indices of the cost
matrix, so the multiset of pairwise costs is identical to the true circumplex but each cost is
assigned to the wrong pair of classes; we draw an independent derangement per seed. The
uniform control replaces every off-diagonal cost with their common mean, which removes all
graded geometry. The uniform case is informative because the loss then collapses to
$\mathcal{L}_{\mathrm{OT}} = \bar{D}\,(1 - p_y)$ with $\bar{D}$ the mean cost, a generic
penalty on the true class's complement: any improvement it produces is attributable to
confidence regularization, not to affect geometry. If the true geometry carries an effect, it
must separate from these controls.

Macro-F1 counts every confusion equally, so it is invariant to which wrong class receives a
prediction's mass and cannot, by construction, measure whether the geometry reshapes the
errors. We therefore score the validation confusion matrix with three geometry-aware
quantities. The error severity is the mean circumplex distance of a misclassification,
$\mathbb{E}[D(\hat y, y) \mid \hat y \neq y]$, macro-averaged over the true class; lower means
the mistakes land on affectively nearer classes. The quadrant preservation is the fraction of
errors whose predicted class shares the true class's valence and arousal sign. The
affect-weighted accuracy is $1 - \mathbb{E}[D(\hat y, y)] / D_{\max}$, a score on which a
near-miss costs less than a far-miss. All three are scored under the true circumplex even for
the control models, and all are reported with paired tests of the geometry against the uniform
control.

\Cref{alg:protocol} states the resulting protocol. We offer it as a template: any
structured-cost loss for affect, and more broadly any graded label cost, can be subjected to
the same shuffled and uniform controls and the same count-versus-geometry comparison, which is
what distinguishes a geometric effect from a generic one.

\begin{algorithm}[t]
\caption{Controlled evaluation of a structured-cost loss}
\label{alg:protocol}
\begin{algorithmic}[1]
\State estimate per-class centroids $\mu_c$ from training VA; set $D_{cc'}=\lVert\mu_c-\mu_{c'}\rVert_2$
\State $D^{\mathrm{unif}}\!\gets$ off-diagonal mean of $D$ (diagonal $0$) \Comment{geometry-free}
\For{seed $s = 1 \dots S$}
  \State $D^{\mathrm{perm}}_s \gets P_s\,D\,P_s^{\!\top}$, $P_s$ a random derangement \Comment{distances kept, geometry scrambled}
  \For{$\hat D \in \{D,\ D^{\mathrm{perm}}_s,\ D^{\mathrm{unif}}\}$ and base (no OT)}
    \State train the identical recipe with $\mathcal{L}_{\mathrm{OT}}$ using $\hat D$; record val confusion
  \EndFor
\EndFor
\State \textbf{score test:} paired permutation test of macro-F1 / $P_{\mathrm{MTL}}$, true cost vs.\ uniform
\State \textbf{geometry test:} same test on error severity and quadrant preservation (scored under $D$)
\State \Return geometric effect $\iff$ true cost is tied with uniform on the score yet beats it on the geometry-aware metrics
\end{algorithmic}
\end{algorithm}

\section{Experimental Setup}
\label{sec:exp:datasets}
All experiments share one recipe; the only difference between compared conditions is the
EXPR loss. Models train on a single 16\,GB consumer GPU in under 25 minutes per run, and we
report mean$\pm$std over seeds with exact paired permutation tests across matched seeds (the
smallest achievable two-sided $p$ at $n{=}6$ is $0.031$). We use three benchmarks.
Aff-Wild2 (\cref{sec:protocol}), the corpus of the ABAW challenges, is the primary
multi-task benchmark; its circumplex is estimated locally as in \cref{sec:method:geometry}. AffectNet~\cite{mollahosseini2017affectnet}
provides eight expression categories with continuous valence--arousal, so the same local
estimation applies and the term enters the identical two-task (EXPR$+$VA) instance of the
framework; we use the official split with the standard eight categories (about $414$k
training and $5.5$k validation images after dropping the non-expression categories), and the
AU mask is empty throughout. RAF-DB~\cite{li2017rafdb} provides categorical expression labels
only ($12{,}271$ training and $3{,}068$ test aligned images over seven basic expressions),
with no valence--arousal annotation; its circumplex therefore cannot be estimated, and it
serves as the geometry-transfer test of \cref{sec:results:cross}, importing the AffectNet
geometry under a fixed mapping of RAF-DB's categories into the AffectNet label order. Since
that mapping leaves one AffectNet class (contempt) without a RAF-DB counterpart, RAF-DB
macro-F1 is reported over its seven present classes. For the two image datasets each image is
its own temporal group, so the leakage check reduces to the official split disjointness, which
the released partitions provide.

A note on the statistics. With six seeds, the paired permutation test has $2^6=64$ sign
assignments, so its smallest attainable two-sided $p$ is $2/64=0.031$; a result reported at
$p=0.031$ therefore means the paired difference is positive (or negative) in all six seeds,
a statement about sign-consistency rather than effect size. We treat the two designed primary
contrasts (the uniform control versus the circumplex on the score, and the action-unit-routing
test) as pre-planned single hypotheses. For the secondary error-structure family
(\cref{sec:results:errstruct}) we additionally apply a Holm correction and report percentile
bootstrap confidence intervals over the six seeds, so that floor-level $p$-values are not read
as graded confidence. All numbers come from logged result files via deterministic scripts.

\section{Results}
\label{sec:results}

\subsection{Influence of the Backbone Pretraining Domain}
\label{sec:results:backbone}
\begin{figure}[t]
  \centering
  \includegraphics[width=0.92\linewidth]{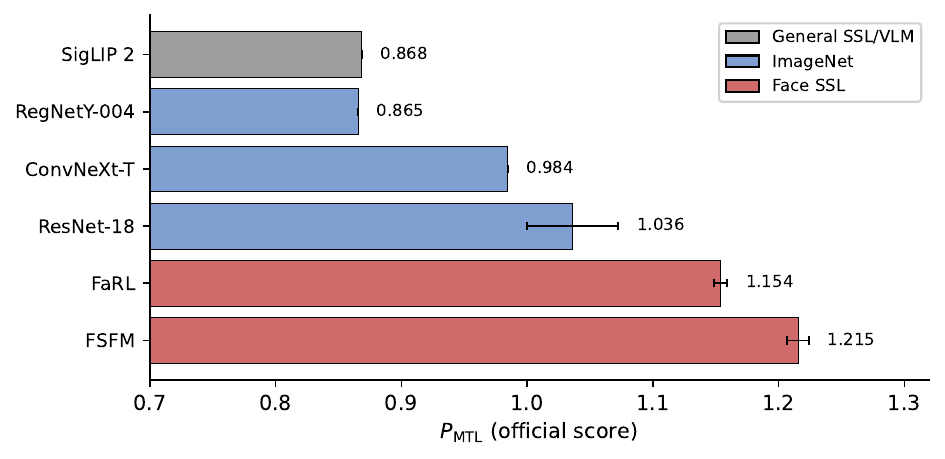}
  \caption{Backbone comparison under an identical recipe on Aff-Wild2.}
  \label{fig:backbone}
\end{figure}

Before introducing the transport term we fix the base encoder by a controlled comparison:
identical recipe, data, and schedule, varying only the pretrained
backbone. The outcome is monotone in how face-specialized the pretraining is
(\cref{fig:backbone}): a 2025 general vision--language encoder (SigLIP\,2,
$0.868$)~\cite{tschannen2025siglip2} trails even ImageNet CNNs; the face vision--language
encoder FaRL~\cite{zheng2022farl} reaches $1.154\pm0.005$; and the face-SSL encoder
FSFM~\cite{wang2025fsfm} leads at $1.215\pm0.009$. Capacity and recency are not the levers;
the pretraining domain is. All subsequent experiments use the FSFM encoder; FaRL serves as a
permissively licensed alternative and, in \cref{sec:degeneracy:rsa}, as a contrasting
representation whose pretraining already encodes affect geometry.

\subsection{The Score Gain Is Real but Not Geometric}
\label{sec:results:main}
\Cref{tab:main} shows the controlled comparison on Aff-Wild2. The OT term improves the
official score from $1.215\pm0.009$ to $1.228\pm0.008$ and EXPR macro-F1 from $0.288\pm0.008$
to $0.297\pm0.007$, with the paired difference positive in all six seeds and significant for
both metrics ($p=0.031$), at no cost to VA or AU. Taken alone, this looks like a successful
geometric method. The geometry
controls of \cref{sec:method:controls} show it is not what it appears.

\begin{table}[t]
  \centering
  \caption{Circumplex-OT versus the base recipe on Aff-Wild2 (mean$\pm$std over six seeds, identical recipe and backbone).} 
  \label{tab:main}
  \resizebox{\columnwidth}{!}{%
  \begin{tabular}{lcccc}
    \toprule
    Method & $P_{\mathrm{MTL}}\uparrow$ & VA (CCC)$\uparrow$ & EXPR (F1)$\uparrow$ & AU (F1)$\uparrow$ \\
    \midrule
    Base recipe (focal CE) & \underline{1.215 $\pm$ 0.009} & \underline{0.438 $\pm$ 0.009} & \underline{0.288 $\pm$ 0.008} & \textbf{0.489 $\pm$ 0.003} \\
    + Circumplex-OT (ours) & \textbf{1.228 $\pm$ 0.008} & \textbf{0.442 $\pm$ 0.006} & \textbf{0.297 $\pm$ 0.007} & \underline{0.489 $\pm$ 0.002} \\
    \bottomrule
  \end{tabular}}
\end{table}

\Cref{tab:shuffled} repeats the comparison with the cost matrix replaced. The uniform cost,
which strips the term of all geometry and reduces it to a confidence penalty, matches the true
circumplex: $P_{\mathrm{MTL}} = \valCtrlUniPmtl{}$ against the circumplex's $1.228$, a
difference that is not significant ($p=\valCtrlCircUniP{}$), and its EXPR macro-F1 is if
anything higher. The permuted cost, which preserves every pairwise distance but assigns it to
the wrong class pair, recovers most of the gain as well. Both controls point to the same
conclusion: the score improvement is the effect of adding any non-trivial optimal-transport
penalty, not of the affect geometry.

\begin{table}[t]
  \centering
  \caption{Geometry control on Aff-Wild2 (six seeds).} 
  \label{tab:shuffled}\resizebox{\columnwidth}{!}{%
  \begin{tabular}{lcccc}
    \toprule
    Ground cost & $P_{\mathrm{MTL}}$ & EXPR (F1) & $p$ vs base & $p$ vs ours \\
    \midrule
    Base recipe (no OT) & 1.215 $\pm$ 0.009 & 0.288 $\pm$ 0.008 & -- & 0.031 \\
    Circumplex cost (ours) & 1.228 $\pm$ 0.008 & 0.297 $\pm$ 0.007 & 0.031 & -- \\
    Permuted geometry & 1.224 $\pm$ 0.007 & 0.295 $\pm$ 0.008 & 0.031 & 0.125 \\
    Uniform cost & 1.227 $\pm$ 0.008 & 0.299 $\pm$ 0.010 & 0.031 & 0.625 \\
    \bottomrule
  \end{tabular}}
\end{table}

The full system context, reported for completeness, does not change this reading. With
inference-time additions (per-AU thresholds calibrated on validation and Gaussian temporal
smoothing within each video) the single model reaches $1.267\pm0.009$, and a cross-backbone
ensemble of the six face-SSL and three vision--language models reaches $1.296$;
\cref{tab:prior} places these numbers among published ABAW MTL systems for context, noting
that editions differ in data release and protocol. These are properties of the recipe, not of
the circumplex term, which the controls have just shown to be geometry-agnostic at the score
level.
\begin{table}[t]
  \centering
  \caption{Comparision with prior works on ABAW MTL validation.}
  \label{tab:prior}
  \footnotesize
  \setlength{\tabcolsep}{4pt}
  \begin{tabular}{lcc}
    \toprule
    System  & Setting & $P_{\mathrm{MTL}}$ \\
    \midrule
    VGG-FACE baseline~\cite{kollias2024abaw7}  & single & 0.32 \\
    AU-relation graph~\cite{abaw2022augraph} & single & 1.288 \\
    Task-adaptive AU graph~\cite{abaw2024auassisted} & single & 1.254 \\
    HSEmotion~\cite{savchenko2024hsemotion}  & blend & 1.494 \\
    Progressive learning~\cite{abaw2024progressive} & multi-stage & 1.791 \\
    \midrule
    Ours, single (raw) & single & 1.228 \\
    Ours, single (+cal.+smooth)  & single & 1.267 \\
    Ours, ensemble & 9 models & 1.296 \\
    \bottomrule
  \end{tabular}
\end{table}

\subsection{Replication of the Generic Gain on a Second Benchmark}
\label{sec:results:cross}
A single benchmark cannot distinguish a property of the method from a property of the
dataset, so we repeat the comparison on AffectNet, where the circumplex is again estimated
locally from the dataset's own valence--arousal labels (\cref{tab:generalize}). The transport
term improves expression macro-F1 from \valAnBaseExpr{} to \valAnCircExpr{}
(\valAnExprDelta{}) over six seeds, roughly six times the Aff-Wild2 gain, in line with the
larger headroom of the AffectNet baseline. The larger gain does not rescue the geometric
interpretation. The uniform control reaches EXPR macro-F1 \valCtrlAnUniExpr{}
(\valCtrlAnUniVsBase{} over base), which not only matches but significantly exceeds the
circumplex ($p=\valCtrlAnLinVsUniP{}$ for the circumplex below the uniform cost). On both
benchmarks, then, the score gain is generic: a geometry-free confidence penalty does at least
as well as the affect geometry, and on the dataset with more headroom it does better. We
include RAF-DB, which carries no valence--arousal annotation and so cannot estimate its own
circumplex, only to note that importing the AffectNet geometry as a fixed external cost yields
no improvement either (\valRafDelta{}, $p=0.125$); a transported constant is no more
geometric than a uniform one.

\begin{table*}[t]
  \centering
  \caption{The circumplex-OT term versus base across three benchmarks.
  }
  \label{tab:generalize}
  \begin{tabular}{lllccccc}
    \toprule
    Dataset & Seeds & Geometry & Base EXPR & +Circ-OT EXPR & $\Delta$ & $p$ & VA (CCC) \\
    \midrule
    Aff-Wild2 & 6 & locally estimated & \underline{0.288 $\pm$ 0.008} & \textbf{0.297 $\pm$ 0.007} & $+0.009$ & 0.0312 & 0.438 / 0.442 \\
    AffectNet-8 & 6 & locally estimated & \underline{0.507 $\pm$ 0.003} & \textbf{0.560 $\pm$ 0.002} & $+0.052$ & 0.0312 & 0.660 / 0.665 \\
    RAF-DB (basic) & 6 & imported (transfer) & \textbf{0.810 $\pm$ 0.008} & \underline{0.805 $\pm$ 0.008} & $-0.005$ & 0.1250 & -- \\
    \bottomrule
  \end{tabular}
\end{table*}

\begin{figure}[t]
  \centering
  \includegraphics[width=\linewidth]{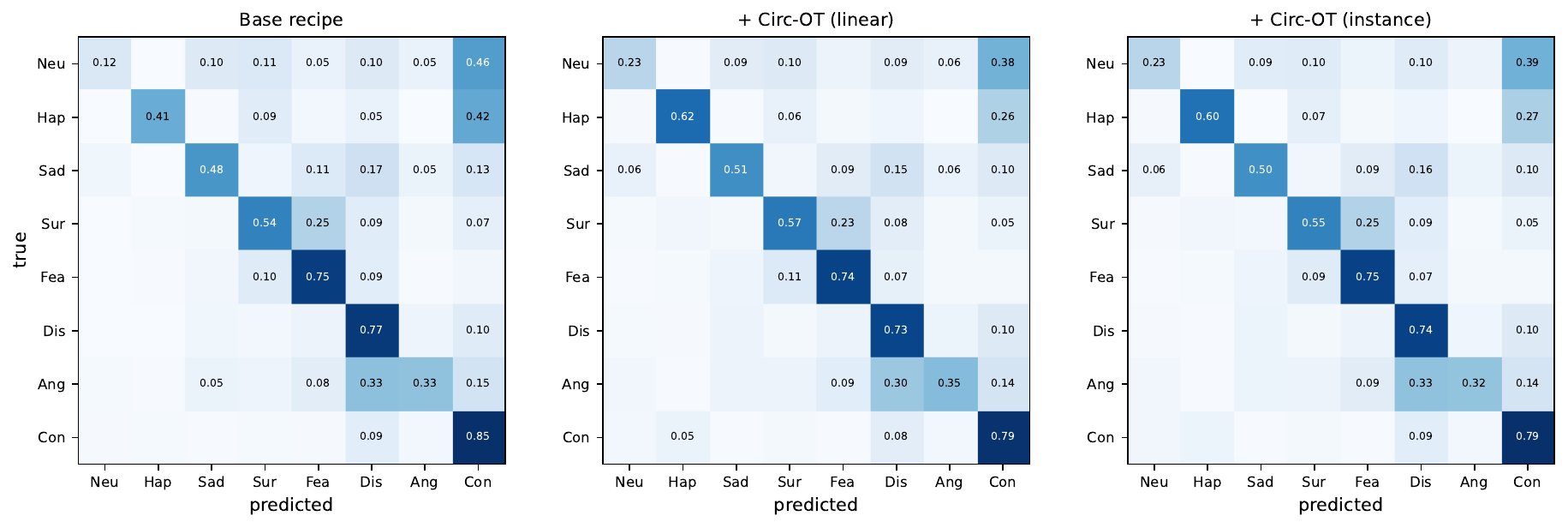}
  \caption{Row-normalized EXPR confusion on AffectNet (validation, seed 0): base, linear
  circumplex-OT, and instance-adaptive variant.
  }
  \label{fig:confusion_an}
\end{figure}

The one place the two datasets genuinely differ is in their degeneracy, which is where the
diagnosis of \cref{sec:degeneracy} begins. Each dataset's circumplex has its own co-located
pair, anger--fear on Aff-Wild2 ($d=0.054$) and anger--contempt on AffectNet ($d=0.064$), and
in both cases the transport term leaves that pair untouched (\cref{fig:confusion_an}): the
anger$\to$contempt confusion on AffectNet barely moves, exactly as anger$\to$fear does not
move on Aff-Wild2. The degeneracy is a stable, dataset-specific fact about affect geometry,
and it is the part of the picture the cost cannot change.

\subsection{Ablation and Dose--Response Analysis}
\label{sec:results:ablation}
\begin{figure}[t]
  \centering
  \begin{minipage}[t]{0.49\linewidth}
    \centering
    \includegraphics[width=\linewidth]{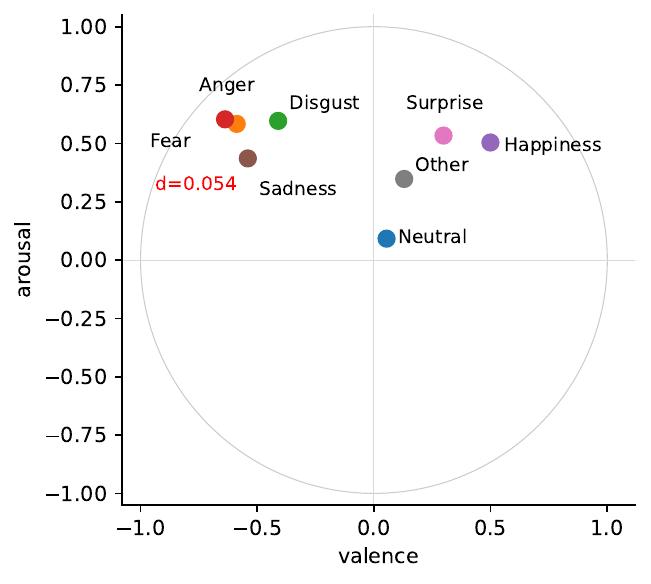}
    \caption{Empirical Aff-Wild2 circumplex: per-class mean (valence, arousal). Anger and
    fear are nearly co-located ($d{=}0.054$; maximum distance $1.142$).}
    \label{fig:circumplex}
  \end{minipage}\hfill
  \begin{minipage}[t]{0.49\linewidth}
    \centering
    \includegraphics[width=\linewidth]{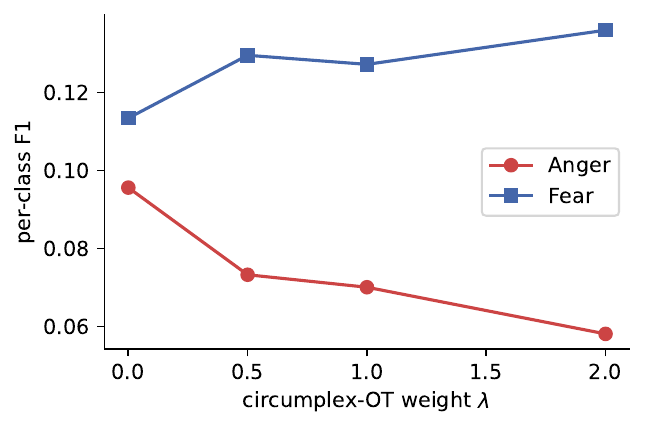}
    \caption{Dose--response of the anger--fear pair under the pure transport weight $\lambda$
    (seed 0). Mass shifts inside the co-located pair while the overall score rises.}
    \label{fig:dose}
  \end{minipage}
\end{figure}

\Cref{tab:ablation} separates the two candidate mechanisms on Aff-Wild2 at seed 0. The OT
ground-cost term alone achieves the full gain with no VA cost. The consistency term alone
scores below the baseline, since it drags VA toward class centroids and destroys within-class
VA variation; adding it to the OT term only dilutes the effect. Increasing the OT weight to
$\lambda{=}2$ yields no further improvement, and halving it retains most of the gain. A sweep of the
pure transport weight (\cref{fig:dose}) shows the term's signature behavior inside the
crowded region of the circumplex: as $\lambda$ grows, anger F1 falls while fear F1 rises and
the overall score increases. Mass migrates to exactly where the ground cost says confusions
are cheap.

\begin{table}[t]
  \centering
  \caption{Term and weight ablation (seed 0). The OT ground-cost term carries the effect; the explicit EXPR$\leftrightarrow$VA consistency term is harmful and is dropped from the final method.}
  \label{tab:ablation}
  \resizebox{\columnwidth}{!}{%
  \begin{tabular}{lcccc}
    \toprule
    Method & $P_{\mathrm{MTL}}\uparrow$ & VA (CCC)$\uparrow$ & EXPR (F1)$\uparrow$ & AU (F1)$\uparrow$ \\
    \midrule
    Base recipe & 1.202 & 0.433 & 0.280 & \underline{0.488} \\
    OT only ($\lambda{=}1$) & \textbf{1.216} & \textbf{0.439} & \textbf{0.290} & 0.487 \\
    Consistency only & 1.199 & 0.424 & 0.287 & \textbf{0.489} \\
    Both & 1.210 & 0.432 & 0.290 & 0.488 \\
    OT ($\lambda{=}2$) + cons. & 1.210 & 0.433 & 0.289 & 0.487 \\
    OT ($\lambda{=}0.5$) + cons. & \underline{1.212} & \underline{0.434} & \underline{0.290} & 0.488 \\
    \bottomrule
  \end{tabular}}
\end{table}

\subsection{Distributional Variants of the Ground Cost}
\label{sec:results:distributional}
\Cref{tab:distributional} evaluates the two distributional refinements of
\cref{sec:method:distributional} against the linear form on Aff-Wild2. The instance-adaptive
cost, which anchors each frame's transport cost at its own annotated valence--arousal point,
matches the linear form on expression macro-F1 and moves $P_{\mathrm{MTL}}$ from
\valDistLinPmtl{} to \valDistInstPmtl{}, a difference inside noise ($p=\valDistInstP{}$
against the linear form over six matched seeds). The dispersion-aware Gaussian-Wasserstein
cost of \cref{eq:w2}, pre-registered as a single seed, returns essentially to the baseline
(\valDistGwPmtl{}): letting within-class covariance reshape the cost does not relax the
anger--fear degeneracy, because the two classes are co-located in their means and overlap in
their dispersion as well. The pattern repeats on AffectNet, where the linear form also edges
the instance variant. The conclusion is consistent with the rest of the paper: the
simplest instantiation of the circumplex cost is as good as its richer ones, and the natural
distributional elaborations, while well motivated, do not extract more from this geometry on
these data. This removes the distributional formulation as a confound before we turn to the
controls.

\begin{table}[t]
  \centering
  \caption{Distributional refinements of the cost on Aff-Wild2.
  }
  \label{tab:distributional}
  \resizebox{\columnwidth}{!}{%
  \begin{tabular}{lcccc}
    \toprule
    Ground cost & Seeds & $P_{\mathrm{MTL}}$ & EXPR (F1) & $p$ \\
    \midrule
    Base recipe (no OT) & 6 & 1.215 $\pm$ 0.009 & 0.288 $\pm$ 0.008 & -- \\
    Linear centroid cost (ours) & 6 & 1.228 $\pm$ 0.008 & 0.297 $\pm$ 0.007 & 0.031 \\
    Instance-adaptive cost & 6 & 1.232 $\pm$ 0.006 & 0.297 $\pm$ 0.007 & 0.094 \\
    Gaussian-Wasserstein cost & 1 & 1.216 & 0.289 & -- \\
    \bottomrule
  \end{tabular}}
\end{table}

\subsection{Comparison with Imbalance-Oriented Methods}
\label{sec:results:altloss}
\Cref{tab:altloss} places the term against standard loss-side treatments of the long tail on
the identical pipeline: margin and asymmetric losses (LDAM~\cite{cao2019ldam} for EXPR with
ASL for AU) and PCGrad gradient surgery~\cite{yu2020pcgrad}. None improves the overall score,
and none separates the anger--fear pair. The transport term does raise the score where these
do not, but the controls of \cref{sec:results:main} attribute that to its confidence-penalty
component rather than to geometry; what the transport term shares with every imbalance method
here is that none of them separates the degenerate pair, the first sign of the diagnosis we
develop in \cref{sec:degeneracy}.

\begin{table}[t]
  \centering
  \caption{Comparison with alternative loss-side interventions (seed 0).
  }
  \label{tab:altloss}
  \begin{tabular}{lcccc}
    \toprule
    Intervention & $P_{\mathrm{MTL}}$ & EXPR F1 & Anger & Fear \\
    \midrule
    Focal CE (base) & 1.2020 & 0.280 & 0.096 & 0.113 \\
    LDAM + ASL & 1.1849 & 0.285 & 0.120 & 0.081 \\
    PCGrad (grad.\ surgery) & 1.1985 & 0.266 & 0.049 & 0.066 \\
    Circumplex-OT (ours) & 1.2163 & 0.290 & 0.070 & 0.127 \\
    \bottomrule
  \end{tabular}
\end{table}

\subsection{Per-Class Analysis}
\label{sec:results:perclass}
\Cref{tab:perclass} breaks the Aff-Wild2 EXPR gain down by class, and \cref{tab:an_perclass} does
the same on AffectNet. On both datasets the gain is a broad reshaping of the error structure
rather than the lift of a single class. On AffectNet the largest improvements fall on the
high-frequency but high-confusion classes (neutral and happiness, each above $+0.15$), while
the geometrically co-located anger--contempt pair ($d=0.064$) is left almost untouched,
mirroring the anger--fear behavior on Aff-Wild2. The pattern is the same on both corpora: the term
helps where the geometry can separate classes and is inert where it cannot. On the AU side the
intervention is surgical (\cref{tab:perau}): every one of the twelve per-unit F1 scores is
identical to the base recipe within rounding and AU macro-F1 is unchanged, confirming that the
EXPR-side term reshapes the expression posterior and touches nothing else.

\begin{table}[t]
  \centering
  \caption{Per-class expression F1 (mean\,$\pm$\,std over six seeds). The circumplex-OT term lifts most classes; inside the co-located anger--fear pair it trades anger down for fear up (\cref{sec:degeneracy:au}).}
  \label{tab:perclass}
  \begin{tabular}{lccc}
    \toprule
    Class & Base recipe & + Circumplex-OT & $\Delta$ \\
    \midrule
    Neutral & 0.388\,$\pm$\,0.031 & 0.414\,$\pm$\,0.015 & +0.026 \\
    Anger & 0.107\,$\pm$\,0.037 & 0.079\,$\pm$\,0.024 & -0.028 \\
    Disgust & 0.444\,$\pm$\,0.031 & 0.449\,$\pm$\,0.007 & +0.006 \\
    Fear & 0.113\,$\pm$\,0.027 & 0.132\,$\pm$\,0.014 & +0.018 \\
    Happiness & 0.443\,$\pm$\,0.013 & 0.440\,$\pm$\,0.005 & -0.003 \\
    Sadness & 0.300\,$\pm$\,0.014 & 0.324\,$\pm$\,0.015 & +0.024 \\
    Surprise & 0.180\,$\pm$\,0.009 & 0.174\,$\pm$\,0.014 & -0.006 \\
    Other & 0.329\,$\pm$\,0.029 & 0.362\,$\pm$\,0.012 & +0.034 \\
    \bottomrule
  \end{tabular}
\end{table}

\begin{table}[t]
  \centering
  \caption{Per-class EXPR F1 on AffectNet-8 (mean over six seeds), base versus the linear circumplex-OT term.
  }
  \label{tab:an_perclass}
  \begin{tabular}{lccc}
    \toprule
    Class & Base & +Circ-OT & $\Delta$ \\
    \midrule
    Neutral & \underline{0.193} & \textbf{0.345} & $+0.151$ \\
    Happiness & \underline{0.560} & \textbf{0.716} & $+0.156$ \\
    Sadness & \underline{0.563} & \textbf{0.593} & $+0.029$ \\
    Surprise & \underline{0.557} & \textbf{0.586} & $+0.029$ \\
    Fear & \underline{0.647} & \textbf{0.656} & $+0.009$ \\
    Disgust & \underline{0.570} & \textbf{0.577} & $+0.007$ \\
    Anger & \underline{0.437} & \textbf{0.443} & $+0.007$ \\
    Contempt & \underline{0.531} & \textbf{0.563} & $+0.032$ \\
    \bottomrule
  \end{tabular}
\end{table}

\begin{table}[t]
  \centering
  \caption{Per-AU F1 on Aff-Wild2 (mean over six seeds), base versus the circumplex-OT term.
  }
  \label{tab:perau}
  \begin{tabular}{lcclcc}
    \toprule
    AU & Base & +OT & AU & Base & +OT \\
    \midrule
    AU1 & 0.578 & 0.577 & AU12 & 0.661 & 0.662 \\
    AU2 & 0.361 & 0.361 & AU15 & 0.171 & 0.171 \\
    AU4 & 0.569 & 0.571 & AU23 & 0.169 & 0.166 \\
    AU6 & 0.561 & 0.558 & AU24 & 0.167 & 0.169 \\
    AU7 & 0.730 & 0.728 & AU25 & 0.830 & 0.829 \\
    AU10 & 0.719 & 0.719 & AU26 & 0.354 & 0.354 \\
    \bottomrule
  \end{tabular}
\end{table}

\subsection{What the Geometry Does Change Is the Error Structure}
\label{sec:results:errstruct}
\begin{figure}[t]
  \centering
  \includegraphics[width=0.92\linewidth]{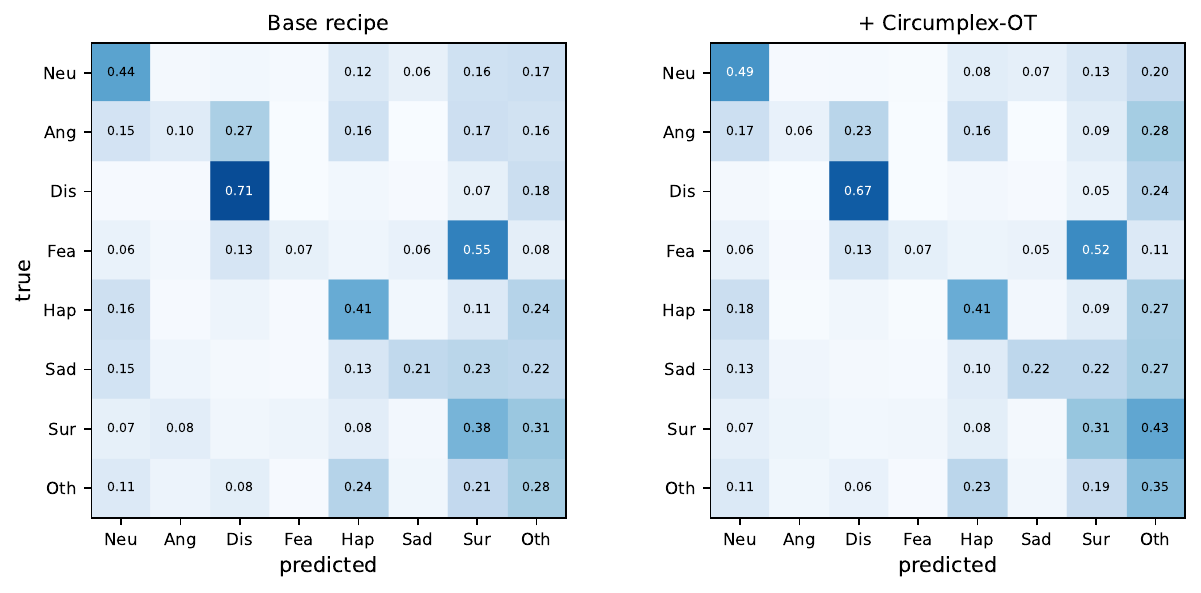}
  \caption{Row-normalized EXPR confusion (Aff-Wild2 validation, seed 0), without and with the
  circumplex-OT term.
  }
  \label{fig:confusion}
\end{figure}

The controls leave one possibility open. Macro-F1 is invariant to which wrong class receives a
prediction's mass, so the circumplex could reshape the model's errors without moving the
score, an effect the uniform control would not reproduce and macro-F1 could not detect. We
test this directly with the error-structure metrics of \cref{sec:method:controls}, comparing
the circumplex against the score-matched uniform control (\cref{tab:errstruct}). On Aff-Wild2
the two are tied on macro-F1, as they must be, but the circumplex makes the surviving errors
affectively nearer the truth (error severity lower, $p=\valEsAbawSevP{}$, bootstrap mean
$\valCiAbSevMean{}$, 95\% CI $\valCiAbSevCI{}$) and more often quadrant-preserving
($p=\valEsAbawQuadP{}$); the score difference, by contrast, has a bootstrap CI that includes
zero ($\valCiAbScoreMean{}$, $\valCiAbScoreCI{}$), consistent with the null on the score. We
do not over-read these two effects, however: each is individually at the $n{=}6$ floor and
neither survives a Holm correction across the four error-structure tests ($\valHolmReject{}$ of
$\valHolmFamily{}$ rejected at $\alpha{=}0.05$), so we report the error-shaping as suggestive
rather than established. The confusion matrices (\cref{fig:confusion}) illustrate the proposed
mechanism: under the term, anger's affectively distant confusion with surprise roughly halves
while mass consolidates in its circumplex neighborhood.

\begin{table}[t]
  \centering
  \caption{Circumplex-OT minus the score-matched uniform control, paired permutation $p$ over six seeds.
  }
  \label{tab:errstruct}
  \begin{tabular}{lcccccc}
    \toprule
    & \multicolumn{2}{c}{EXPR-F1} & \multicolumn{2}{c}{Error severity} & \multicolumn{2}{c}{Quadrant keep} \\
    \cmidrule(lr){2-3}\cmidrule(lr){4-5}\cmidrule(lr){6-7}
    Dataset & $\Delta$ & $p$ & $\Delta$ & $p$ & $\Delta$ & $p$ \\
    \midrule
    Aff-Wild2 & $-0.002$ & 0.406 & $-0.008$ & 0.031 & $+0.011$ & 0.031 \\
    AffectNet & $-0.005$ & 0.031 & $-0.002$ & 0.094 & $+0.004$ & 0.156 \\
    \bottomrule
  \end{tabular}
  \\[2pt]
  {\footnotesize $\Delta$ is circumplex-OT minus uniform cost; negative error severity and positive quadrant keep favor the geometry.}
\end{table}

This error-shaping does not survive on AffectNet (\cref{tab:errstruct}, lower row): there the
circumplex is tied with the uniform control on error severity ($p=\valEsAnSevP{}$) and on
quadrant preservation, and relative to the base recipe it makes errors slightly farther rather
than nearer. The cause is a single confusion, contempt mistaken for happiness, the farthest
pair in the AffectNet geometry: contempt's unilateral smile is visually close to happiness
even though the two are affectively opposite, so the cost shaping competes with appearance
structure and loses. The straightforward conclusion is that the geometry-specific effect is suggestive
where we can see it and absent on the other dataset, because visual confounds can overwhelm it.
The score gain is generic on both benchmarks; the error effect is, at best, geometric on one
and absent on the other. Neither is the durable finding of this study. That finding is the
limit itself.

\section{The Rare-Class Limit Is a Circumplex Degeneracy}
\label{sec:degeneracy}
Across every experiment above, one fact does not move: the model fails on the same rare
classes, and no cost we apply rescues them. This section assembles the evidence that the
failure is a geometric degeneracy of the affect space, representational in origin, and beyond
the reach of loss or cost design.

\subsection{Representational Alignment and the Backbone}
\label{sec:degeneracy:rsa}
The transport term reshapes errors on the face self-supervised encoder but is inert on the
face vision--language encoder, and the reason is representational. We test the hypothesis that
vision--language pretraining already places affectively similar expressions near one another
in feature space, so the circumplex cost adds structure the representation already has. For
each encoder state we compute class-prototype features on the validation set,
take their pairwise cosine-distance matrix, and correlate it (Spearman over the 28 class
pairs) with the circumplex distance matrix (\Cref{tab:rsa}). The frozen
FaRL representation is already aligned with the circumplex ($\rho=\valRsaFarlFrozen{}$),
whereas the frozen FSFM representation is not ($\rho=\valRsaFsfmFrozen{}$); training FSFM with
the term pushes its alignment upward
(\valRsaBaseMean{}$\pm$\valRsaBaseStd{} at the base recipe to
\valRsaCircMean{}$\pm$\valRsaCircStd{} with the term, over six seeds), though the shift is
small and not significant (paired permutation $p=\valRsaShiftP{}$); the load-bearing contrast
is instead the frozen one, where face vision--language pretraining is already aligned and pure
face self-supervision is not. The conditionality is representational: the term has
label-similarity structure to supply only where pretraining has not, and is redundant where
language supervision has already supplied it. Alignment is necessary but
not sufficient for recognition, however, and one row of \cref{tab:rsa} makes this plain: the
general vision--language encoder SigLIP\,2 is the most circumplex-aligned of all
($\rho=\valRsaSiglip{}$) yet the weakest backbone of \cref{sec:results:backbone}. Encoding the
affect geometry does not by itself guarantee discriminability; face-domain pretraining adds the
class separability that a general vision--language model, however well aligned, lacks. The
reinterpretation of the backbone study is therefore narrower than a clean dichotomy: face
vision--language features carry affect geometry that leaves the term little to add, but
geometry alone does not explain the backbone ranking.

\begin{table}[t]
  \centering
  \caption{Alignment between encoder feature geometry and the circumplex.
  }
  \label{tab:rsa}
  \begin{tabular}{lc}
    \toprule
    Encoder / state & $\rho$(feature, circumplex) \\
    \midrule
    FSFM pretrained (frozen) & $+0.071$ \\
    FSFM fine-tuned (base) & $+0.151$ \\
    FSFM fine-tuned (+circ-OT) & $+0.179$ \\
    FaRL pretrained (frozen) & $+0.233$ \\
    FaRL fine-tuned (base) & $+0.244$ \\
    FaRL fine-tuned (+circ-OT) & $+0.250$ \\
    SigLIP2 fine-tuned & $+0.369$ \\
    ResNet-18 fine-tuned & $+0.090$ \\
    \bottomrule
  \end{tabular}
\end{table}

\subsection{Persistence of the Degenerate Pair under Richer Geometry}
\label{sec:degeneracy:au}
If the rare-class failure were a property of the circumplex cost specifically, a better cost
should fix it. Action units offer the obvious better cost, since anger and fear have distinct
prototypical configurations (brow lowering versus brow raising, lip tightening versus mouth
stretching) even though they nearly coincide in valence and arousal. We estimate a per-class
action-unit signature, the mean AU vector over the frames carrying both an expression and AU
labels, and form a composite ground cost that averages the circumplex distance with the
AU-signature distance rescaled to the same units. This lifts the anger--fear pair from a
near-degenerate $0.054$ to a well-resolved $0.335$, so a cost on this geometry can, in
principle, separate them.

\Cref{tab:negatives} reports the outcome on Aff-Wild2 over six seeds. Anger F1 does not
recover: with the VA-only cost it is \valAngCirc{} against the base recipe's \valAngBase{},
and with the enriched VA$+$AU cost it is \valAngAU{}, no better and significantly below base
either way, while overall EXPR is unchanged from the VA-only cost. Making the ground cost
separate the pair changes nothing about the model's ability to separate it. This is the
cleanest statement of the diagnosis: the collapse of anger is not a property of the circumplex
cost, since a cost built to resolve the pair does not help, but of the representation, which
cannot tell anger from its neighbors regardless of how any cost weights the confusion. A cost
can reweight confusion among classes the features already separate; it cannot create
separability the features lack.

\begin{table}[t]
  \centering
  \caption{AU-routing on Aff-Wild2 (six seeds).
  }
  \label{tab:negatives}
  \resizebox{\columnwidth}{!}{%
  \begin{tabular}{lccc}
    \toprule
    Cost geometry & EXPR (F1) & Anger (F1) & $p$ vs base \\
    \midrule
    Base recipe & 0.288 $\pm$ 0.008 & 0.107 $\pm$ 0.037 & -- \\
    + Circ-OT (VA cost) & 0.297 $\pm$ 0.007 & 0.079 $\pm$ 0.024 & 0.031 \\
    + Circ-OT (VA+AU cost) & 0.297 $\pm$ 0.007 & 0.078 $\pm$ 0.029 & 0.031 \\
    \bottomrule
  \end{tabular}}
\end{table}

\subsection{Class Isolation and Class Reachability}
\label{sec:degeneracy:isolation}
The degeneracy view suggests a quantitative pattern: how much a class can be helped at all
should track how isolated it is in affect space, because a class with a distant nearest
neighbor has affect-distant confusions to shed while a co-located class has none that any cost
can remove. We test this by correlating each class's isolation, the distance to its nearest
neighbor on the circumplex, with its change in F1 across seeds. The rank correlation is
positive on both benchmarks ($\rho=\valIsoAbaw{}$ on Aff-Wild2, $\rho=\valIsoAffectnet{}$ on
AffectNet), though neither is significant alone (two-sided permutation $p=\valIsoAbawPtwo{}$ and
$\valIsoAnPtwo{}$); combining the two datasets gives $p=\valIsoCombinedP{}$, one-sided in the
direction the mechanism predicts and stated at analysis time rather than pre-registered, and
the eight classes per dataset are not fully independent since the degenerate pairs share an
isolation value. We therefore interpret this as consistent with the degeneracy account, not as an
established predictor. The one robust observation is that the single class to lose F1 anywhere
is anger, the most co-located class on either circumplex, and this holds whether the help comes
from the circumplex cost or, as the controls show, from a generic penalty: reachability is a
property of the representation and the affect geometry, not of the particular cost.

\section{Discussion and Limitations}
\label{sec:discussion}
The methodological lesson of this study is that a plausible geometric improvement can be
entirely generic, and only a control distinguishes the two. The circumplex-cost term improves
the multi-task score on two benchmarks, yet a uniform cost, which carries no geometry,
reproduces the improvement and on AffectNet exceeds it. Without the control, the natural and
incorrect conclusion is that affect geometry helps; the control attributes the gain to the
confidence regularization that any non-trivial transport penalty supplies. We suggest that
shuffled and uniform-cost controls become standard practice for structured-cost losses in
affect recognition, where graded label costs are increasingly common and rarely controlled.

The geometry is not entirely ineffective: it reshapes the error structure, making mistakes
affectively nearer the truth on Aff-Wild2. The effect is 
suggestive rather than established, individually at the six-seed significance floor and not
surviving a Holm correction across the error-structure family, and it does not reproduce on
AffectNet, where a single visually grounded confusion, contempt mistaken for happiness,
overwhelms it. The honest reading is that affect-geometry supervision may shape error quality
where visual structure does not contradict it, but cannot be relied on to do so across
datasets. Establishing where the effect holds, and whether a cost that also models appearance
could stabilize it, is open. More broadly, our conclusions rest on validation splits of two
locally-estimable corpora and a single visual modality, with no public held-out test set; the
generality of the diagnosis is bounded accordingly, and ``across datasets'' here means the two we
examine, not a population. A distributional formulation with an entropic coupling did not
improve on the linear cost in our preliminary experiments.

The finding that does hold across every control and both datasets is the diagnosis. The rare
classes fail because they are degenerate on the circumplex, and that degeneracy is
representational: frequency-based losses do not fix it, the transport term does not, and a
cost built specifically to separate anger from fear does not. Class isolation, computed from
the geometry before training, is consistent with which classes are reachable. The implication
is constructive. Progress on anger and fear will not come from reweighting or repricing their
confusions but from representations that separate them. The action-unit channel, which
distinguishes the pair in principle even though routing it through the cost did not help,
remains the most promising source of orthogonal evidence, to be fused at the feature level
rather than the loss.

\section{Conclusion}
\label{sec:conclusion}
We aimed to use the geometry of affect as a training signal and, through controls that this
literature rarely runs, found that its apparent benefit to the multi-task score is a generic
regularization effect rather than a geometric one, while its genuine geometric effect, on the
structure of the errors, is real on one benchmark and absent on another. What survives every
control is a diagnosis: the persistent failure of rare expressions in in-the-wild affect
recognition is a circumplex degeneracy that is representational in origin, beyond the reach of
any loss or cost, and resolvable only by features that tell the classes apart. The controls
and error-structure metrics we introduce are a contribution in their own right: they are what
separated a real effect from a convincing artifact.

\bibliographystyle{IEEEtran}
\bibliography{refs}

\end{document}